# Bidirectional LSTM-CRF Attention-based Model for Chinese Word Segmentation

Chen Jin[a,*], Zhuangwei Shi[b,*], Weihua Li[c], Yanbu Guo[c]

*a. College of Computer Science, Nankai University, Tianjin, P.R.China*

*b. College of Artificial Intelligence, Nankai University, Tianjin, P.R.China*

*c. Department of Information Science and Engineering, Yunnan University, Kunming, Yunnan, P.R China*

*\*Correspondence: {jinchen_cs, zwshi}@mail.nankai.edu.cn*

**Abstract**

Chinese word segmentation (CWS) is the basic of Chinese natural language processing (NLP). The quality of word segmentation will directly affect the rest of NLP tasks. Recently, with the artificial intelligence tide rising a gain, Long Short-Term Memory (LSTM) neural network, as one of easily modeling in sequence, has been widely utilized in various kinds of NLP tasks, and functions well. Attention mechanism is an ingenious method to solve the memory compression problem on LSTM. Furthermove, inspired by the powerful abilities of bidirectional LSTM models for modeling sequence and CRF model for decoding, we propose a Bidirectional LSTM-CRF Attention-based Model in this paper. Experiments on PKU and MSRA benchmark datasets show that our model performs better than the baseline methods modeling by other neural network.

## 1. Introduction

Chinese word segmentation (CWS) is a fundamental task for Chinese natural language processing. Unlike English or other western languages, Chinese does not have a blank to separate words. Therefore, word segmentation is a preliminary and significant pre-process for Chinese language processing. In recent years, Chinese word segmentation has been greatly developed. The most popular method is to treat this task as a sequence labeling problem [26] [21]. The goal of sequence labeling is to set correct labels to all the words in a sequence, which can be dealt with supervised learning algorithms such as Hidden Markov Model(HMM) [20], Maximum Entropy (ME) [1] and Conditional Random Fields (CRF) [15]. However, these models are restricted by the pre-designing features. Moreover, the number of features could be large and most of them are always useless. So that the result models are too large for practical use and prone to overfit on training corpus. Recently, neural network models have wildly been used in various NLP tasks. Due to this, we no longer need to pick suitable features by hand. Collobert et al. [7] proposed a general neural network architecture for sequence labeling tasks, and Zheng et al. [28] applied the architecture to Chinese word segmentation and POS tagging. Following these works, different kinds of neural network have been applied to Chinese word segmentation task and achieved impressive performance. Among these works, Recurrent Neural Network (RNN) especially Long short-Term Memory (LSTM) [12] neural network and its improved model, achieved the outstanding improvement. Chen et al [4] firstly used LSTM Neural Network for Chinese word segmentation task. Chen et al. [3] proposed a gated recursive neural network (GRNN) which is a kind of tree structure to capture long distance dependencies. Huang et al. [13] combined a Bidirectional LSTM network and a CRF network to form a Bi-LSTM-CRF network. Peng and Dredze [22] applied the architecture to Chinese word segmentation and named entity recognition (NER).

However, there is still a problem existing in classical LSTM network model. The issue relates to memory compression problems. As the input sequence gets compressed and blended into a single dense vector, sufficiently large memory capacity is required to store past information. As a result, the network generalizes poorly to long sequences while wasting memory on shorter ones. Through the improvement of LSTM unit with attention mechanism, the Long Short-Term Memory-Network (LSTM) [6] is proposed to solve the problem.

In this paper, we propose a bidirectional LSTM-CRF attention-based model, and apply it to Chinese word segmentation task. Inspired by the success of Bi-LSTM-CRF model and LSTM unit, we replace LSTM unit to traditional LSTM unit in Bi-LSTM-CRF Model. We also use the layer-wise training method to avoid the problem of gradient diffusion, and the dropout strategy to avoid the overfitting problem.

The contributions of this paper can be summarized as follows:

- We propose a bidirectional LSTM-CRF attention-based model, use attention mechanism to solve memory compression problems.

- Our model can be easily generalized and applied to the other sequence labeling tasks, such as Part-of-speech tagging(POS), chunking and named entity recognition (NER).

- The performance of our model is evaluated by Chinese word segmentation on PKU and MSRA benchmark datasets which are commonly used for evaluation of Chinese word segmentation. Experiment results show that our model performs better than the baseline methods modeling by other neural network.

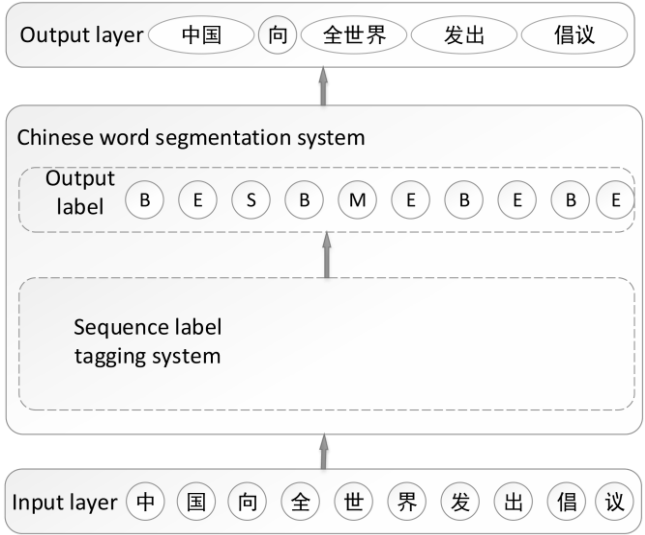

Figure 1: Structure of Long Short-Term Memory Network for Chinese segmentation.

The remainder of the paper is organized as follows. Section 2 describes how to transform Chinese word segmentation task to a sequence labeling problem. Section 3 proposes Bidirectional LSTM-CRF Attention-based Model. Section 4 describes the model we proposed for Chinese word segmentation task and training procedure. Section 5 reports the experiments results. The related research is discussed in Section 6 . Finally we conclude in Section 7.

Chinese word segmentation task Chinese word segmentation task is usually regarded as a character-based sequence labeling problem. Each character is labeled as one of $\{B, M, E, S\}$ to indicate the segmentation. $B, M, E$ represent **Be**tively, and S represents a **Single** character segmentation. For emample, if we input a general sentence sequence "中国向全世界发出倡议" to a segmentation system, then we will get an output label sequence *BESBMEBEBE*, so that we can transform it to 中国—向—全世界—发出—倡议.

**2. Bidirectional** LSTM**-CRF Neural Networks**

*2.1. LSTM Networks with Attention Mechanism*

Long Short-Term Memory (LSTM) neural network [12] is an extension of the Recurrent Neural network (RNN). It has been designed to use a memory-cell to capture long-range dependencies.

LSTMs can produce a list of state representations during composition, however, the next state is always computed from the current state. While the recursive state update is performed in a Markov manner, it is assumed that LSTMs maintain unbounded memory this assumption may fail in practice, for example when the sequence is long or when the memory size is not large enough.

To address this limitation. Cheng [6] modified the standard LSTM structure by replacing the memory cell with a memory network [25]. The resulting Long Short-Term Memory-Network (LSTM) stores the contextual representation of each input token with a unique memory slot and the size of the memory grows with time until an upper bound of the memory span is reached. This design enables the LSTM to reason about relations between tokens with a neural attention layer and then perform non-Markov state updates.

The architecture of the LSTM is shown in Figure 1 and the formal definition is provided as follows. The model maintains two sets of vectors stored in a hidden state tape used to inter-act with the environment, and a memory tape used to represent what is actually stored in memory. Therefore, each token is associated with a hidden vector and a memory vector. Let $x_t$ denote the current input; $C_{t-1} = (c_1, ..., c_{t-1})$ denote the current memory tape, and $H_{t-1} = (h_1, ..., h_{t-1})$ the previous hidden tape. At time step t, the model computes the relation between $x_t$ and $x_1, ..., x_{t-1}$ through $h_1, ..., h_{t-1}$ with an attention layer:

$$a_i^t = v^T tanh(W_h h_i + W_x x_t + W_{\tilde{h}} \tilde{h}_{t-1}) \quad (1)$$

$$s_i^t = softmax(a_i^t) \quad (2)$$

This yields a probability distribution over the hidden state vectors of previous tokens. We can then compute an adaptive summary vector for the previous hidden tape and memory tape denoted by $\tilde{c}_t$ and $\tilde{h}_t$, respectively:

$$\begin{bmatrix} \tilde{h}_t \\ \tilde{c}_t \end{bmatrix} = \sum_{i=1}^{t-1} s_i^t \cdot \begin{bmatrix} h_i \\ c_i \end{bmatrix} \quad (3)$$

and use them for computing the values of $c_t$ and $h_t$ in the recurrent update as:

$$\begin{bmatrix} i_t \\ f_t \\ o_t \\ \hat{c}_t \end{bmatrix} = \begin{bmatrix} \sigma \\ \sigma \\ \sigma \\ tanh \end{bmatrix} W \cdot [\tilde{h}_t, x_t] \quad (4)$$

$$c_t = f_t \odot \tilde{c}_t + i_t \odot \hat{c}_t \quad (5)$$

$$h_t = o_t \odot tanh(c_t) \quad (6)$$

where $v$, $W_h$, $W_x$ and $W_{\tilde{h}}$ are the new weight terms of the network. A key idea behind the LSTM is to use attention for inducing relations between tokens. These relations are soft and differentiable, and components of a larger representation learning network.

*2.2. Bidirectional LSTM-CRF Networks*

The bidirecitional LSTM neural network using conditional random fields (CRF)[15] as the output interface for sentence-level optimisation (Bi-LSTM-RNN) achieve state-of-the-art accuracies on various sequence tagging tasks [13] [17] [16]

One shortcoming of conventional LSTMs networks is that they are only able to make use of previous context. In segmentation task, we always need both forward and backward information in context. Bidirectional LSTM networks [24] do this by processing the data in both directions.

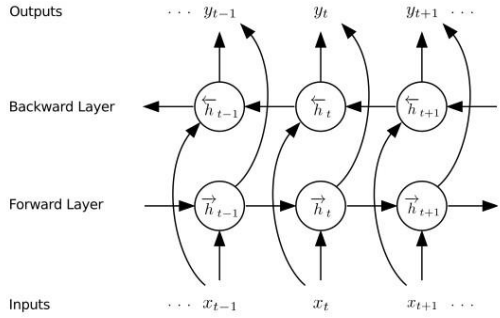

Figure 2: Structure of Bidirectional LSTM unit.

An LSTM computes the forward hidden sequence $h$, the back-ward hidden sequence $h$ and the output sequence $y$ by iterating the backward layer from $t = T$ to 1, the forward layer from $t = 1$ to $T$ and then updating the output layer:

$$\overrightarrow{h}_t = H(W_{x\overrightarrow{h}}x_t + W_{\overrightarrow{h}\overrightarrow{h}}\overrightarrow{h}_{t-1} + b_{\overrightarrow{h}}) \quad (7)$$

$$\overleftarrow{h}_t = H(W_{x\overleftarrow{h}}x_t + W_{\overleftarrow{h}\overleftarrow{h}}\overleftarrow{h}_{t+1} + b_{\overleftarrow{h}}) \quad (8)$$

$$y_t = W_{\overrightarrow{h}y}\overrightarrow{h}_t + W_{\overleftarrow{h}y}\overleftarrow{h}_t + b_y \quad (9)$$

We train the bidirectional LSTM networks model using back-propagation through time [2]. The forward and backward passes over the unfolded network over time are carried out in a similar way to regular network forward and backward passes, except that we need to unfold the hidden states for all time steps. We also need a special treatment at the beginning and the end of the data points. In our implementation, we do forward and backward for whole sentences and we only need to reset the hidden states to 0 at the begging of each sentence. We have batch implementation which enables multiple sentences to be processed at the same time.

We find that it yields a dramatic improvement over single-layer LSTM.

For sequence labeling (or general structured prediction) tasks, there has been the correlations between labels in neighborhoods and jointly decode the best chain of labels [7] for a given input sentence.

In Chinese segentation $\{B, M, E, S\}$ system, for example, label $B$ is followed by label $M$ or $E$ instead of label $S$. And label $M$ is followed by label $M$ or $E$ rather than label $S$ or $B$.

Therefore, instead of modeling tagging decisions independently, we add a linear chain conditional random field [15] layer as the decoder.

Formally, we use $x = (x_1, ..., x_n)$ to represent an input sequence, and use $y = (y_1, ..., y_n)$ to represent the segmentation output label. So, we define its scores to be

$$s(x_{1:n}, y_{1:n}, \theta) = \sum_{i=0}^{n} A_{y_i,y_{i+1}} + \sum_{i=0}^{n} P_{\theta,i,y_i} \quad (10)$$

where $A$ is a matrix of transition score such that $A_{i,j}$ to model the transition from the $i$-th state to $i$-th state for pair of consecutive time steps. $y_0$ and $y_n$ are the start and end tags of a sentence, that we add to the set of possible tags. $P$ is the matrix of scores output by the bidirectional LSTM netural network. The parameter sets of netural network is $\theta$.

A softmax over all possible tag sequences yields a probability for the sequence $y$:

$$p(y|x) = \frac{e^{s(x,y,\theta)}}{\sum_{\tilde{y} \in y_x} e^{s(x,\tilde{y},\theta)}} \quad (11)$$

where the $y_x$ denotes the set of possible label sequences for $x$.

During training, we use maximum likelihood estimation to maximize the log-probability of the correct tag sequence:

$$ln(p(y|x)) = s(x, y, \theta) - ln(\sum_{\tilde{y} \in y_x} e^{s(x,\tilde{y},\theta)}) \quad (12)$$

where $y_x$ represents all possible tag sequences for a sentence $x$. From the formulation above, it is evident that we encourage our network to produce a valid sequence of output labels.

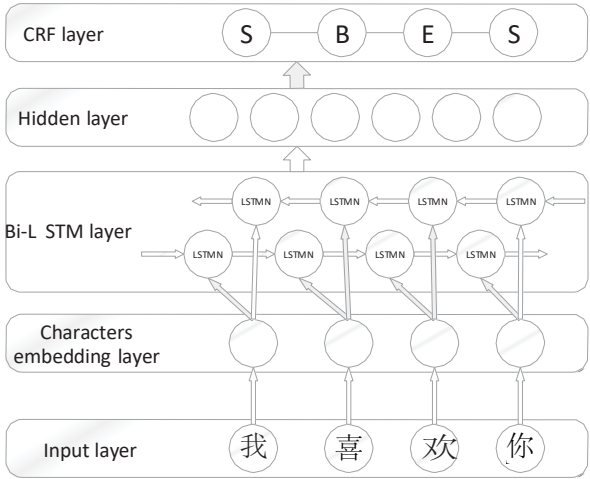

Figure 3: Structure of Long Short-Term Memory Network for Chinese segmen-tation.

While decoding, we predict the output sequence that obtains the maximum score $y^*$ given by:

$$y_* = \underset{\tilde{y}\in y_x}{\operatorname{argmax}} s(x,\tilde{y},\theta) \qquad (13)$$

Training and decoding can be solved efficiently by adopting the dynamic programming(Viterbi algorithm) [23] [15].

### 2.3. Bidirectional LSTM-CRF Attention-based Neural Networks

We replace the classical LSTM unit with LSTM unit in Bidirectional LSTM neural networks to form a new nework architecture. Beside this, we combine this new network and a CRF network to form the bidirectional LSTM-CRF Attention-based neural networks (a illustration of the model architecture is shown in Figure 2), that can use various of models' advantages.

## 3. Model training Procedure

### 3.1. Model

The LSTM neural network architecture for Chinese word segmentation task is characterized by three specialized layers: (1) a character embedding layer; (2) bidirectional LSTM neural network layer and (3) CRF tag decoder layer. A illustration is shown in Figure 3.

**Character Embeddings Layer** Firstly, we need to transform the input sentence to vector data, when we use neural network model to dispose data. Previous studies have shown that distributed representation is superior to one-hot representation for modeling sequences, such as sentences or genes [18][10]. Word2Vec [19] is a classical method for learning distributed representation for sequences. In this paper, we use Word2Vec to get the character embed-ding as the top layer of this model.

**Bidirectional LSTM neural network layer** The embeddings of all the context characters are concatenated into a single vector as input of the forward LSTM neural network as well as the backward LSTM neural network. Then, it will go through a series hidden layer,and get a matrix of scores.

**CRF tag decoder layer** In this layer, we define a transition score join to the matrix of scores from the neural network as evaluation score. Using the chain conditional random field modeling the input sequence and output label. The label that got maximum evaluation score is the predicted label from system.

### 3.2. Training Algorithm

For the models presented in this paper, we train our models using mini-batched AdaGrad [8]. In each epoch, we divide the whole training data to batches and process one batch at a time. Each batch contains a list of sentences which is determined by the parameter of mini-batch size. In our experiments, we use mini-batch size of 50. For each mini-batch, we first run bidirectional LSTM-CRF model forward pass which includes the forward pass for both forward state and backward state of LSTM. As a result, we get the the output score for all tags at all positions. We then run CRF layer forward and backward pass to compute gradients for network output and state transi-tion edges. After that, we can back propagate the errors from the output to the input, which includes the backward pass for both forward and backward states of LSTM. Finally we up-date the network parameters which include the state transition matrix $A$, and the original bidirectional LSTM parameters.

### 3.3. Overfitting

Dropout is one of prevalent methods to avoid overfitting in neural networks [8] [11]. When dropping a unit out, we temporarily remove it from the network, along with all its incoming and outgoing connections. In the simplest case, each unit is omitted with a fixed probability $p$ independent of other units, namely dropout rate, where $p$ is also chosen on development set.

## 4. Experiment

### 4.1. Dataset and Setup

We evaluate our proposed approach on three datasets, SogouCA, PKU and MSRA. SogouCA is provided by the So-gouLab. The PKU and MSRA data both are provided by the second International Chinese Word Segmentation Bakeoff [9]. We mixed PKU's, MSRA's training dataset and SogouCA as a dataset to pre-trained the character embedding. Using PKU and MSRA to do the Segmentation. As a general, we randomly divide the whole training data into the 90% sentences as training set and the rest 10% sentences as development set.

| Embedding Size | P | R | F |
|---|---|---|---|
| 50 | 95.5 | 95.1 | 95.3 |
| **100** | **96.6** | **96.3** | **96.4** |
| 150 | 96.1 | 95.6 | 95.8 |

Table 1: performance of Bi-LSTM-CRF Attention-base model with different embedding dimensions

All datasets are preprocessed by replacing the Chinese idioms and the continuous English characters and digits with a unique flag.

We use Tensorflow 0.12 to build the neural network for training and testing, and use Word2Vec toolkit to pre-train the dataset to input vector data. It takes about 13 hours (not inlcude the pre-computation or parameter loading time) to train the model, and training procedure speeded up by GPU (GTX 1060 6G).

For evaluation, we use the standard bake-off scoring program to calculate precision, recall, F1-score.

*4.2. Hyper-parameters*

Hyper-parameters of neural model impacts the performance of the algorithm significantly. According to experiment results, we choose the hyperparameters of our model as follow. The minibatch size is set to 50. Generally, the number of hidden units has a limited impact on the performance as long as it is large enough. We found that 150 with one layer is a good trade-off between speed and model performance. The dimensionality of character embedding is set to 100 which achieved the best performance. All these hyperparameters are chosen according to their average performances on development sets.

*4.3. Experiment Result*

In this section, we will state the procedure of our experiments and how we get the model with the best performance, and we will also compare the performance of our network with other solutions.

Table 1 shows the performances with additional pre-trained and bigram character embeddings. Again, the performances boost significantly as a result. Moreover, when we use bigram embeddings only, which means we do close test without pre-training the embeddings on other extra corpus, our model still perform competitively compared.

Table 2 shows that as we design more hidden layers, the performance gets slight improvement, while adding layers becomes not so effective when the number of hidden layers exceeds is two, which also takes quite long time to train. The result shows that hidden layer become less effective in higher level layers so we believe that there is no need to build very deep network for extracting contextual information.

Table 3 lists the performances of our model as well as previous segmentation systems. CRF++[14] is a CRF word segmentation algorithm, which features are mostly design by hand. While the rest of the list are neural network models, that are vanilla LSTM model [4], and bidirectional-LSTM-CRF model [22]. Our model achieved competitive performance compared with [22], which also used Bi-LSTM-CRF model. Besides, we evaluated the performance per epoch with different dropout rate, and the result was shown on Figure 4.

## 5. Related Work

Chinese word segmentation task is usually regarded as a character-based sequence labeling problem. Recently, neural network models have wildly been used in various of NLP tasks especially in sequence labeling problem. Due to this, we no longer need to pick suitable feature by hand. Collobert et al. [7] proposed a general neural network architecture for sequence labeling tasks, and Zheng et al. [28] applied the architecture to Chinese word segmentation and POS tagging. Following this work, different kind of neural network have been applied to Chinese word segmentation task and achieved impressive performance. Among all of these works, Recurrent Neural Network (RNN) especially Long short-Term Memory (LSTM) [12] Neural Network and its improved model function better. Chen et al. [4] firstly used LSTM Neural Network for Chinese word segmentation task. Chen et al. [3] proposed a gated recursive neural network (GRNN) which is a kind of tree structure to capture long distance dependencies.

Huang et al. [13] combined a Bidirectional LSTM network and a CRF network to form a Bi-LSTM-CRF network. And then, this model has become the most authoritative one, various of model are proposed based on Bi-LSTM-CRF model. Ma and Hovy [17] used some convolutional neural network layer connect to the tail of Bi-LSTM layer, proposed a new Bi-LSTM-CNNS-CRF model. It also performs quite well.

Chen et al. [5] use multi-task learning to train multi-criteria corpus at same time, and utilize the thought of adversarial networks to produce shared features in there model. This successful trial by GANs offer another way to solve this problem.

## 6. Conclusion

In this paper, we propose a bidirectional LSTM-CRF attention-based model and use it as Chinese Word Segmentation tool to train the model for Chinese word segmentation. Bidirectional LSTM-CRF attention-based model is quite efficient for sequential tagging task. The model learns to extract discriminative character-level features automatically and it does not require any handcraft features for segmentation or prior knowledge. Experiments conducted on SIGHAN Backoff 2005 datasets show that our model has good performance. Our results suggest that deep neural networks work well on segmentation tasks and bidirectional LSTM-CRF attention-based with word embedding is an effective tagging solution and worth further exploration such as Part-of-speech tagging(POS), chunking and named entity recognition (NER).

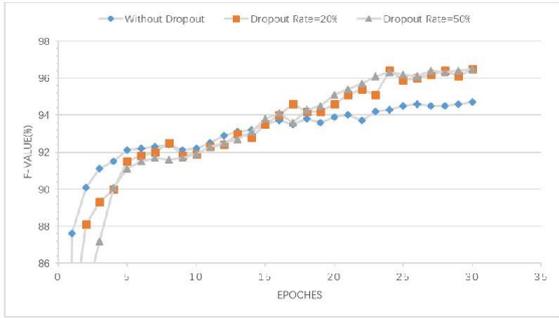 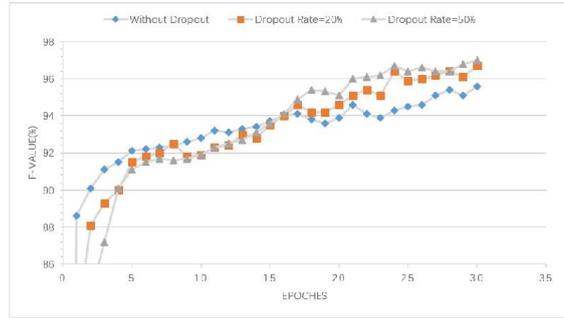

(a)PKU  (b)MRSA

Figure 4: Every epoch preformances of different dropout rate on PKU and MSRA dataset

| Model | MSRA Corpus | | | PKU Corpus | | |
|---|---|---|---|---|---|---|
| | P | R | F1 | P | R | F1 |
| Bi-LSTM-CRF (H *) | 92.6 | 94.0 | 93.3 | 93.6 | 92.1 | 92.8 |
| Bi-LSTM-CRF (H 1) | 96.6 | 96.2 | 96.4 | 95.8 | 95.5 | 95.7 |
| Bi-LSTM-CRF (H 2) | **97.3** | **97.1** | **97.2** | **96.8** | **96.4** | **96.6** |

Table 2: Performance of our models on test sets,Bi-LSTM-CRF (H *)means model without hidden layers,Bi-LSTM-CRF (1 *)means model with 1 hidden layer,Bi-LSTM-CRF (H 2) means model with 2 hidden layers

| Model | MSRA Corpus | | | PKU Corpus | | |
|---|---|---|---|---|---|---|
| | P | R | F1 | P | R | F1 |
| CRF++ [14] | 92.6 | 94.0 | 93.3 | 93.6 | 92.1 | 92.8 |
| LSTM [4] | 96.6 | 96.2 | 96.4 | 95.8 | 95.5 | 95.7 |
| Bi-LSTM | 96.6 | 97.1 | 96.9 | 96.5 | 95.3 | 95.9 |
| Bi-LSTM-CRF | 97.2 | 97.1 | 97.1 | 96.6 | 95.9 | 96.2 |
| **Our Model** | **97.3** | **97.1** | **97.2** | **96.8** | **96.4** | **96.6** |

Table 3: Comparison of our model with previous research